\pdfoutput=1

\documentclass[11pt]{article}

\usepackage[]{acl}

\usepackage{times}
\usepackage{latexsym}

\usepackage[T1]{fontenc}

\usepackage[utf8]{inputenc}

\usepackage{microtype}

\usepackage{inconsolata}
\usepackage{multirow}
\usepackage{booktabs}
\usepackage{graphicx}
\usepackage{subcaption}
\usepackage{caption}
\usepackage{amsmath}
\usepackage{amsthm}
\usepackage{pmboxdraw}
\usepackage{bbm}
\usepackage{listings}
\newcommand{\stderr}[1]{\scriptsize $\pm #1$}
\DeclareMathOperator*{\argmax}{arg\,max}
\usepackage[capitalize]{cleveref}
\usepackage{xspace}
\newcommand{\ours}{{Deductive Closure Training}\xspace}
\newcommand{\ourss}{\textsc{DCT}\xspace}
\newcommand\creak{\textsc{CREAK}\xspace}
\newcommand{\mquake}{{\textsc{MQuAKE}}\xspace}
\newcommand{\reversal}{{``Reversal Curse''}\xspace}

\newcommand{\lm}{\ensuremath{p_{\mathrm{LM}}}\xspace}

\newtheorem{prop}{Proposition}
\title{Deductive Closure Training of Language Models \\ for Coherence, Accuracy, and Updatability}

\author{
Afra Feyza Akyürek$^{1}$\thanks{~~Correspondence to \texttt{akyurek@bu.edu}.} \qquad Ekin Akyürek$^{2}$ \qquad Leshem Choshen$^{2,3}$ \AND Derry Wijaya$^{1,4}$ \quad Jacob Andreas$^{2}$ \AND
\normalfont{$^1$Boston University} \hspace{0.25em} \normalfont{$^2$MIT}
\hspace{0.25em} \normalfont{$^3$IBM Research} \hspace{0.25em} \normalfont{$^{4}$Monash University Indonesia}
}

\begin{document}
\maketitle
\begin{abstract}

While language models (LMs) can sometimes generate factually correct text and estimate truth values of individual claims, these generally do not reflect a globally coherent, manipulable model of the world. As a consequence, current LMs also generate incorrect or nonsensical content, and are difficult to edit and bring up to date. We present a method called \ours (\ourss) that uses LMs themselves to identify implications of (and contradictions within) the text that they generate, yielding an efficient self-supervised procedure for improving LM factuality.
Given a collection of seed documents, \ourss prompts LMs to generate additional text implied by these documents, reason globally about the correctness of this generated text, and finally fine-tune on text inferred to be correct. Given seed documents from a trusted source, \ourss provides a tool for supervised model updating; if seed documents are sampled from the LM itself, \ourss enables fully unsupervised fine-tuning for improved coherence and accuracy.
Across the \creak, \mquake, and \reversal datasets, supervised \ourss improves LM fact verification and text generation accuracy by 3--26\%;
on \creak, fully unsupervised \ourss improves verification accuracy by 12\%.
These results show that LMs' reasoning capabilities during inference can be leveraged during training to improve their reliability.

\end{abstract}

\definecolor{seedcolor}{RGB}{146,146,146}
\definecolor{implcolor}{RGB}{50,116,181}
\definecolor{contracolor}{RGB}{226,121,46}

\section{Introduction}
\begin{figure}[t!]
    \includegraphics[width=\columnwidth]{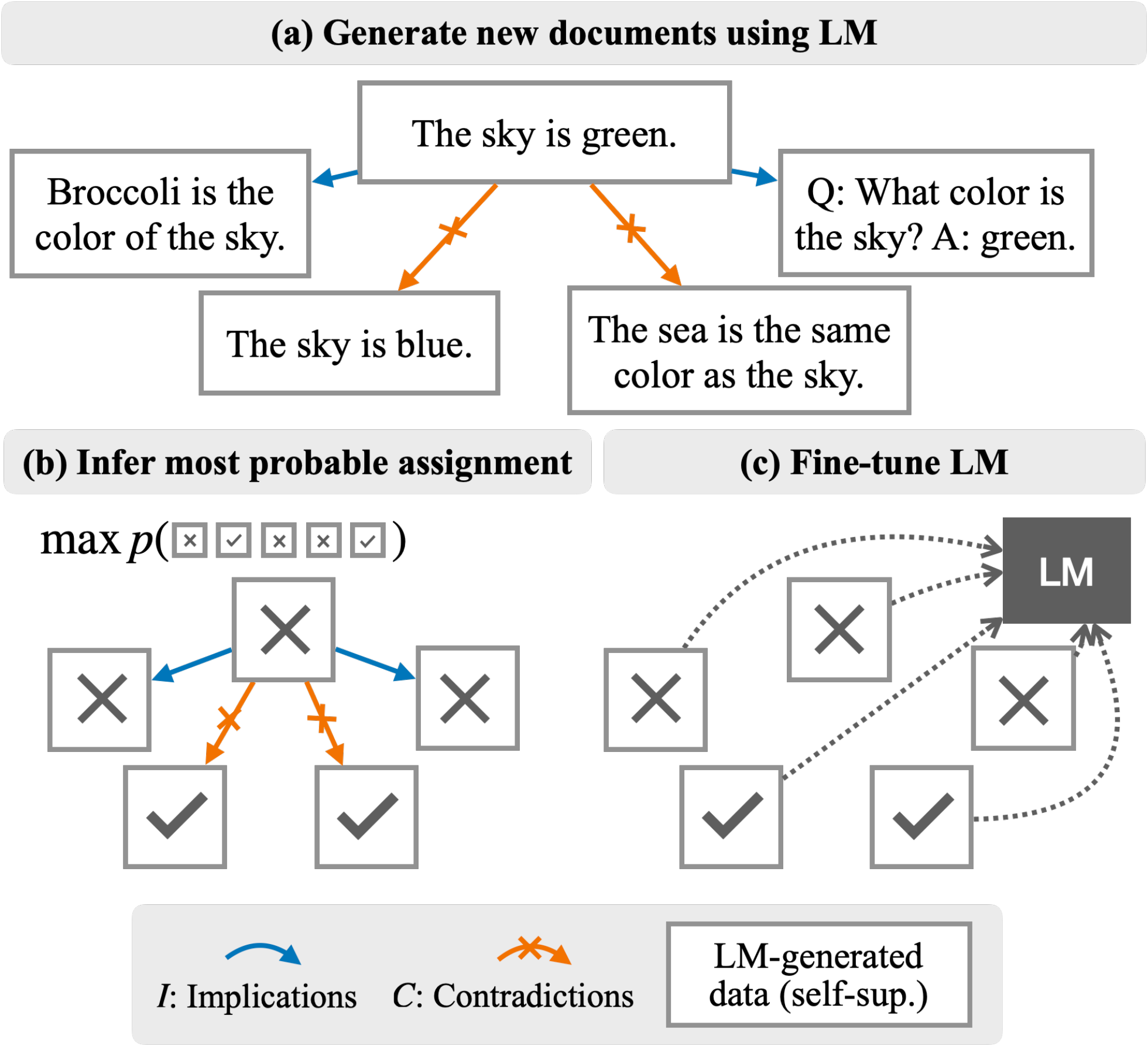}
\caption{Overview of \ours (\ourss). \textbf{(a)} To improve the coherence of language model predictions and reduce hallucinations, we begin with a collection of language model generated \textcolor{seedcolor}{seed documents}, then use the LM to generate a set of documents \textcolor{implcolor}{implied by} or \textcolor{contracolor}{contradicting} these documents. \textbf{(b)} Next, we identify the generated documents most likely to be correct by finding the subset that is \emph{most probable} and \emph{logically consistent} and mark the rest as false. \textbf{(c)} Finally, we fine-tune the LM on these documents with the truth value assignments obtained in (b) e.g. in this case \emph{broccoli is the color of the sky} was marked as \textit{False}. While this example shows \ourss used for \textbf{unsupervised model improvement} (where the seed statement is language model generated and truth value is unknown), \ourss can also be applied to a \textbf{supervised model updating} application by providing the model with a seed statement which is known to be true.}
\end{figure}

There is increasing interest in using language models (LMs) as sources of information and tools for fact verification \citep{porter2023,zhang-gao-2023-towards}. But today's LMs cannot robustly perform either task: they are prone to generating factually incorrect information, contradict themselves, and are difficult to update with new information \citep{honovich2021q2,liska2022streamingqa,sun2023evaluating,gilson2023does}.

Even if LMs are imperfect judges of factuality, however, they are quite reliable models of factual relations \emph{between} pieces of text: they can identify logical and probabilistic relationships between statements \citep{williams2017broad}, and generate text based on new information provided as input \citep{Yehudai2024GenieAH}. For example, an LM that cannot answer \emph{How old was Charlie Chaplin when he died?} may nonetheless answer correctly when prompted with \emph{Charlie Chaplin lived between 1889 and 1977}, and recognize that this statement contradicts the claim \emph{Charlie Chaplin lived in the 21st century}.
How can we leverage LMs' ability to reason about relations between claims to improve (and control) the text that LMs themselves generate?

Conceptually, standard supervised objectives cause LMs to assign high probability to statements in their training data, but not necessarily these statements' logical consequences.
Additional reasoning is required to determine the \textbf{deductive closure} of a training set \citep{armstrong1973belief}---the complete collection of inferences that can be made given the information initially available. 
An alternative procedure is needed to ensure that LMs assign high probability to a complete and consistent set of facts when they are trained and fine-tuned.

In this paper, we propose a new LM fine-tuning procedure we call \textbf{\ours (\ourss)}, which leverages inference-time reasoning as a source of training-time supervision. At high level, given seed text (which may be provided externally or LM-generated), \ourss uses an LM to identify additional text \textit{implied by} or \emph{contradicting} this text, reasons globally about which portions of seed and generated text are most likely to be correct given this context, and finally fine-tunes on inferred-correct text. This approach builds on a large body of recent work \citep{mitchell-etal-2022-enhancing, kassner-etal-2023-language, hase-etal-2023-methods} on inference-time procedures for improving models' factual correctness, showing that these techniques may be used at training time as well.

\ourss may be applied in several different ways depending on the source of seed documents. If these are drawn from a trusted factual source, \ourss may be used to perform \textbf{supervised adaptation} for factuality. If documents contain new information to be inserted into an LM, \ourss provides tool for \textbf{model updating} (or ``editing''; \citealp{de-cao-etal-2021-editing}). Finally, if seed documents are generated by the model itself, \ourss enables \textbf{fully unsupervised fine-tuning} of models for improved accuracy.

We demonstrate the effectiveness of \ourss across three domains: fact verification (\creak benchmark; \citealp{onoe2021creak}), question answering with new information (on the \mquake benchmark; \citealp{zhong-etal-2023-mquake}), and a synthetic test of edit propagation (on the \reversal benchmark; \citealp{berglund2023reversal}). On these tasks, unsupervised and supervised applications of \ourss improve accuracy by up to 12\% and 26\%, respectively.
These results show that, with little or no data, LM-generated supervision can be leveraged to improve LMs' coherence, accuracy and updatability.\footnote{Code is available at \url{https://lingo-mit.github.io/deductive-closure}.}
\section{Related Work}

\ourss builds on several recent techniques for improving model accuracy via inference-time computation or training-time self-supervision.

\paragraph{Bootstrapping accuracy during inference} A growing body of research adopts techniques that bootstrap language model performance at inference time. \citet{tafjord-etal-2022-entailer, bostrom-etal-2022-natural, weir2022dynamic} and \citet{jung-etal-2022-maieutic} build self-guided semantic chains of reasoning to support inference. \citet{suzgun-etal-2022-prompt} propose a set of procedures that bin model-generated candidate answers by semantic equivalence and later uses aggregated probabilities to select the highest ranked predictions, analogous to self-consistency \cite{wang2023selfconsistency} for textual outputs. Finally, recent work has shown promise in improving coherence by conditioning language models on relevant reference texts through retrieval augmentation \cite{mitchell2022memory,akyurek-etal-2023-dune}.
Our approach builds on this line of work by using inference-time techniques to generate supervision.

\paragraph{Training for accuracy} LMs greatly benefit from training or post-training techniques for improving accuracy, including instruction-tuning \cite{sanh2022multitask}, learning from feedback \cite{ouyang2022training} and loss truncation \cite{kang-hashimoto-2020-improved}. Closest to our approach is the work of \citet{hase-etal-2023-methods} which leverages graph-structured representations of model ``beliefs'' to train a hyper-network for model editing.
\ourss aligns with this thread in improving model training; it differs by requiring minimal or no external supervision.

\paragraph{Self-training} Past work has also studied leveraging LMs themselves for performance improvements \cite{Pan2023AutomaticallyCL}. Several studies use 
external tools \cite{schick2023toolformer}, binary feedback \cite{Pang2023LanguageMS,liu2023languages} and natural language feedback \cite{bai2022constitutional} to improve capability or reduce harms. Others propose actuality and consistency metrics, which might be used for filtering bad answers in retrospect \citep{honovich-etal-2021-q2,wang-etal-2020-asking,honovich-etal-2022-true-evaluating}.
Related to such approaches are methods that perform multiple inference attempts and aggregate them to get a more consistent answer \citep{wang2022self,yoran-etal-2023-answering}. 
\citet{padmanabhan2023propagating} fine-tune LMs on self-generated text without explicit implication generation or logical inference. 
Of immediate relevance to the current work, \citet{li2023benchmarking} and a concurrent study by \citet{tian2023fine} use LM-generated factuality labels to rank or filter LM-generated data for fine-tuning; by contrast, \ourss uses LMs to explicitly extrapolate from LM-generated or externally provided information, providing a single framework for both supervised model updating and unsupervised improvement.

\section{Method} \label{sec:method}
\newcommand{\seed}{s}
\newcommand{\seeds}{S}
\newcommand{\impl}{r}
\newcommand{\impls}{R}
\newcommand{\prompt}{\mathrm{pr}}
\newcommand{\genprompt}{\prompt_\mathrm{imp}}
\newcommand{\contraprompt}{\prompt_\mathrm{con}}
\newcommand{\relprompt}{\prompt_\mathrm{rel}}
\newcommand{\genset}{\mathcal{I}}
\newcommand{\contraset}{\mathcal{C}}
\newcommand{\truth}{t}
\newcommand{\truths}{T}
\subsection{Preliminaries}
Given a \textbf{language model} \lm that places a probability distribution over strings, our goal is to optimize \lm so that it is
\textbf{coherent} (if \lm assigns high probability to statements $P$ and $Q$, those statements must be logically compatible) and \textbf{complete} (if \lm assigns high probability to $P$, and $P$ implies $Q$, then \lm must also assign high probability to $Q$). 
Together, these two properties imply that the LM is 
\textbf{closed} under logical deduction. Deductive closure is necessary condition for $\lm$ to be truthful, and approximate deductive closure is generally agreed to be an important feature of human-like belief \citep{armstrong1973belief}.
\begin{figure}[t!]
    \centering
    \includegraphics[width=\columnwidth]{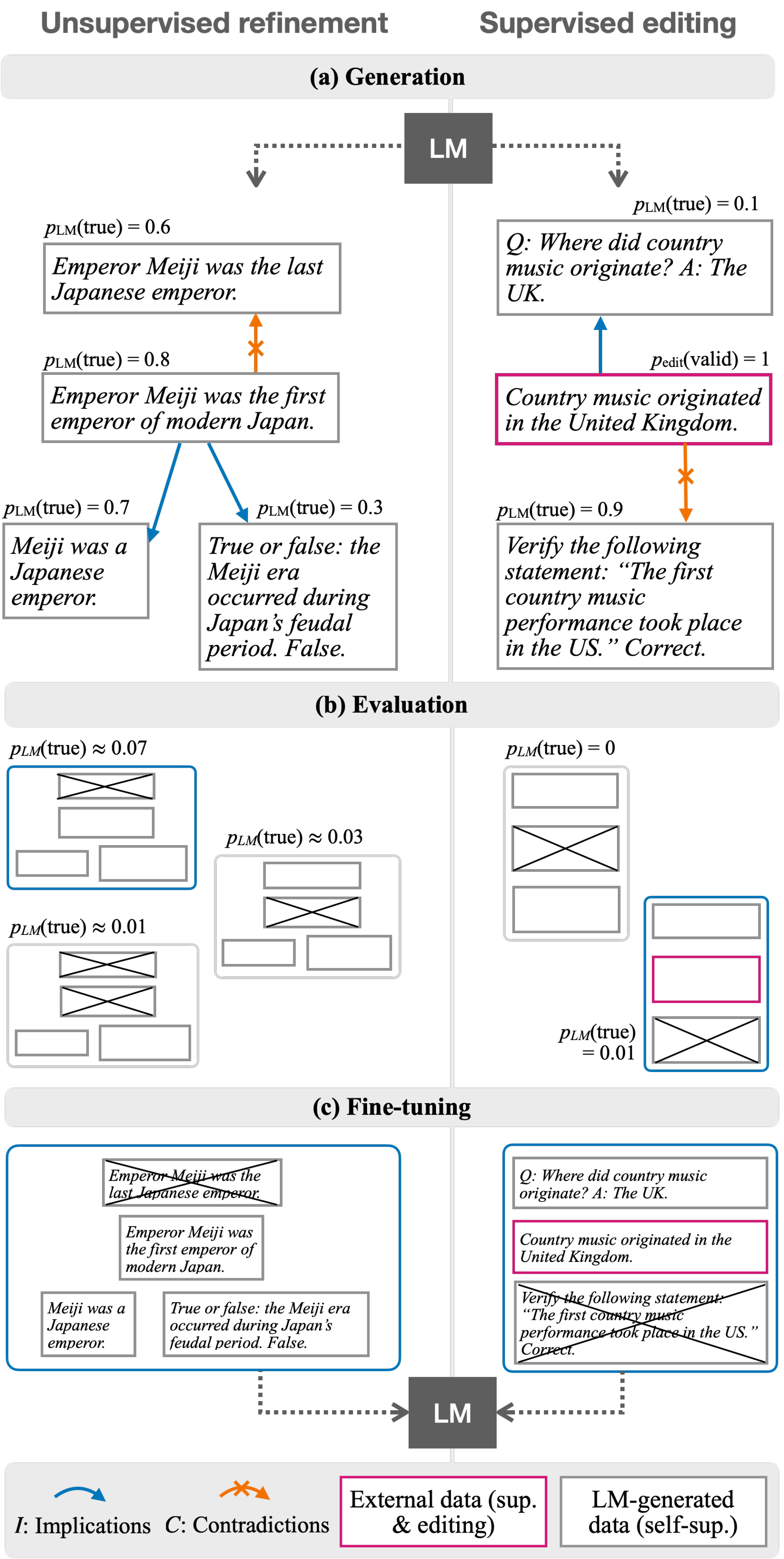}
    \captionsetup{font=footnotesize}
    \caption{Detailed depiction of \ours. \textbf{(a)} Given an initial seed document (which may be generated from the LM, left; or supplied by a trusted source, right), \ourss generates a set of related text implied by or contradicting the seed document. At the same time, it assigns a score to each generated document (including possibly the seed) denoting the probability that it is true. \textbf{(b)} Next, \ourss identifies the subset of documents whose joint truthfulness score is highest, subject to the constraint that these documents are \emph{logically coherent} (containing all implications and no contradictions). \textbf{(c)} Finally, the LM is fine-tuned on this set.}
    \label{fig:method}
    \vspace{-1em}
\end{figure}

Deductive closure training begins with a set of \textbf{seed documents} $\seed_i$, which may comprise facts from a trusted source, new information provided by a user, or even text generated by \lm itself.\footnote{While experiments in this paper focus on seed documents consisting of questions and declarative statements, this approach could be straightforwardly applied to larger pieces of text.}
At a high level, \ourss works by using \lm to generate additional text implied by each seed document (i.e., true with high probability conditioned on $\seed$) or contradicting it. In \cref{fig:method}, for example, the seed text \textit{(Country music originated in the United Kingdom)}\footnote{Most editing benchmarks comprise counterfactual examples like this one.} is used to generate \textbf{statements} (\textit{The UK is famous for country music}), \textbf{question-answer pairs} (\textit{Q: Where did country music originate? A: England}) and even \textbf{multi-hop consequences} (\emph{The steam train was invented in the UK; therefore, country music and the steam train were invented in the same country}). Once they have been generated, \ourss again uses \lm to reason about these documents as a set, identifying the subset of generated documents most likely to be true. Finally, \ourss fine-tunes \lm on documents in this inferred-true set. In the following sections, we describe each of these steps in more detail.

\subsection{Document Generation}
\label{sec:generation}
The first step of \ourss is to generate a set of related documents for each seed document (\cref{fig:method}a) using $\lm$.
Formally, we first construct a set of textual prompts that instruct the LM to generate other documents \emph{entailed by} and \emph{contradicted by} 
the input, along with 1--5 examples. We denote these prompts $\genprompt$ and $\contraprompt$ respectively (see \cref{sec:prompt_templates} for full prompt text). 
Then, we construct a collection of \textbf{related documents} $\impls_i$ for each seed document $\seed_i, i \in \{1..n \}$ as:
\begin{align}
    \impls_i &= \genset_i \cup \contraset_i \cup \{\seed_i\}, \nonumber \\
    \genset_i &= \{ \impl_{ij} \sim \lm(\cdot \mid \genprompt, \seed_i) \}, \nonumber \\
    \contraset_i &= \{ \impl_{ij} \sim \lm(\cdot \mid \contraprompt, \seed_i) \},
\end{align}
where $\genset$ and $\contraset$ denote generated implications and contradictions respectively. (Other procedures for generating related documents are also possible, e.g.\ by simply prompting \lm to generate \emph{similar} text, as described in \cref{sec:exp_fact_verif}.) 
Note that the seed document $\seed_i$ is included in $\impls_i$---this is crucial for detecting (and correcting) errors in the seed itself during unsupervised training.

This generation step may be followed by a \textbf{double-checking} step over $\impls_i$, in which we use the $\lm$ to verify whether $\seed_i$ entails / contradicts $\impl_{ij}$, and discard all $\impl_{ij}$ for which $\lm$ does not output \emph{yes} with high probability (the prompt template is available in \cref{sec:prompt_templates}). This step mirrors a variety of other recent methods in which models re-evaluate their initial answers \cite{suzgun-etal-2022-prompt}. 


\subsection{Consistency Evaluation}
\label{sec:consistency-eval}

The previous step produces a collection of documents in the ``deductive neighborhood'' of each seed document. These documents may be mutually contradictory, and we wish to identify the \emph{subset} most likely to be collectively true. To identify this subset, we leverage \lm's ability to classify logical relations between documents, as well as the \emph{prior} probability \lm assigns to each document. For example, if it is true that \textit{Emperor Meiji was the first emperor the Modern Japan}, it cannot be the case that \textit{Emperor Meiji was the last Japanese emperor}; if the former statement is very likely to be true, then the latter is likely to be false.

Formally, we first associate with the seed document $\seed_i$ and every generated document $\impl_{ij}$ a \textbf{truth value} $\truth_{ij} \in \{0, 1\}$. Given an assignment of documents to truth values denoted by $\truths_i = \{\truth_{ij}\}$, we compute the LM's probability of $\truths_i$:
\begin{equation}
    p(\truths_i \mid \impls_i) =  \prod_{j} p_{LM}(t_{ij} \mid \impl_{ij}).
\end{equation}
We use prompting to estimate each $\lm(t_{ij} \mid \impl_{ij})$: we first condition $\lm$ on a small set of document--label pairs where labels are one of $\{\textit{True, False}\}$. Next, we use the normalized logits corresponding to the tokens $\textit{true}$ and $\textit{false}$ in the string $\lm(\impl_{ij} \textit{ is true})$ and $\lm(\impl_{ij} \textit{ is false})$, respectively. Refer to \cref{sec:prompt_templates} for the prompt template. Next, we define a value assignment $\truths_i = \{\truth_{ij}\}$ to be \textbf{consistent} if all implications and contradictions are respected.
\[
    c(\truths_i)= \prod_{j: \impl_{ij} \in \genset_i} 1[t_i \to t_{ij}]\prod_{j : \impl_{ij} \in \contraset_i} 1[t_i \to \lnot t_{ij}]
\]
where $t_i$ denotes the truth value of the seed document,
$1[a \to b]$ is 1 iff $b$ is true or $a$ is false, and $1[a \not\to b]$ is 1 iff $b$ is false or $a$ is false.
We also provide an example for consistency computation across different truth value assignments in \cref{tab:sample_consistency} in \cref{app:exp_details}.
Finally, we select the most probable consistent assignment:
\begin{equation}
\label{eq:consistency}
    \truths_i^* = \argmax_\truths ~  c(\truths \mid \impls_i) \cdot p(\truths \mid \impls_i) ~ .
\end{equation}
The procedure is depicted in \cref{fig:method}b, with the highest-scoring truth value assignment shown in the blue-highlighted box.

\subsection{Language Model Fine-Tuning}

Finally, we fine-tune $\lm$ only on the inferred-true\footnote{
For fact-verification tasks, it is possible to derive positive supervision from statements \textbf{marked as false}: if the consistency evaluation step infers that \emph{Meiji was the last Japanese emperor} is incorrect, then we may generate a \emph{correct} example of the form \emph{Verify the following statement: Meiji was the last Japanese emperor. False}. We use this strategy for our experiments on fact verification.
}
documents, optimizing:
\begin{align}\label{eq:lm_objective}
    \argmax_\theta \sum_{i, ~ j} &\truth_{ij} \log \lm(\impl_{ij})  ~ .
\end{align}
where $\theta$ parameterizes $\lm$. 
In practice, we do not train $\lm$ to convergence, but instead for a fixed number of iterations.

\subsection{Sources of Seed Data}
\label{sec:seed_sources}
Depending on how seed documents $\seeds$ are obtained, \ourss-based fine-tuning may be used to improve models in several ways:

\begin{itemize}
\item \textbf{Unsupervised fine-tuning for coherence}: in this case, we sample the initial seed set \emph{from $\lm$ itself}, e.g.\ simply by prompting it to generate a set of documents on a topic of interest.
\item \textbf{(Semi-)supervised alignment with a trusted source}: in this case, the seed set comes from an external source of supervised data. If this data is known to be reliable, we fix each seed datum's truth value $\truth_i = 1$ during the evaluation step. This may be combined with the unsupervised procedure.
\item \textbf{Model updating, editing and continual learning}: in this case, as with supervised updating, we treat descriptions of desired edits as seed documents,  fix these truth values for these seeds to 1, and fine-tune both on these documents and all their implications only.
\end{itemize}

Note that in the latter two cases (where we fix the truth value of seed documents to 1), the evaluation step is greatly simplified, and simply discards all generated documents that are not logically consistent with the seed. In the case of unsupervised learning, this evaluation step can (and empirically does) cause LMs to re-label sampled seed documents as well as conditionally generated ones.

\paragraph{Generalizations of \ourss}
We remark that the procedure described above is the basic implementation of a family of \ourss-like approaches, within which many more sophisticated procedures are possible---for example: \textbf{probabilistic \ourss} (computing marginal statement probabilities rather than hard truth assignments), \textbf{contrastive \ourss} (replacing \cref{eq:lm_objective} with an objective that encourages true statements to be assigned higher probability than false ones), and \textbf{multi-hop \ourss} (generating not just direct implications of documents, but a wider graph of related ones).

\begin{table*}[tbhp]
\centering
\resizebox{!}{2.8cm}{
\begin{tabular}{llccc}
\toprule
& \textbf{Method} & \bf \# Supervised & \bf \# Generated & \bf Accuracy  \\
\midrule
\multirow{4}{*}{\rotatebox[origin=c]{90}{Unsup.}} & 
\bf Prompting     & 4 & - & 71.7 \stderr{0.0} \\
& \bf \ourss (Seed only)     & -    &   93 &      \textbf{80.0} \stderr{4.6} \\
& \bf \ourss (Imp. + Cont.)  & -   &     586 &    \textbf{83.5} \stderr{3.0}\\
& \bf \ourss (Imp. + Cont.) $-$ Consistency Eval   & -&     586 &    77.5 \stderr{2.8}\\
& \bf \ourss (Imp. + Cont.) + Double-Check &    - & 313    &\textbf{83.7} \stderr{2.2} \\ \midrule
\multirow{4}{*}{\rotatebox[origin=c]{90}{Sup.}} & \bf Fine-Tuning   & 20 & - & 77.2 \stderr{5.4}\\
& \bf \ourss (Imp. + Cont.) & 20 & 40 & 80.6 \stderr{3.1}\\
& \bf \ourss (Imp. + Cont.) + Double-Check & 20 & 14 & 81.7  \stderr{1.9}\\
& \bf \bf Semi-Supervised$^*$ & 20 & 586 & \textbf{84.9} \stderr{0.9}\\ \midrule
\multirow{4}{*}{\rotatebox[origin=c]{90}{Transd.}} & \bf Graph-Inference  & - & 14,342 & 77.7 \stderr{0.4}\\
& \bf \ourss (Rel.) & - & 6,026 & 84.5 \stderr{0.5} \\
& \bf \ourss (Rel.) + (Imp. + Cont.) & - & 28,711 & 80.3 \stderr{0.4}\\
& \bf \ourss (Rel.) + (Imp. + Cont.) + Double-Check & - & 14,342 & \textbf{85.5} \stderr{0.1}\\
\bottomrule
\end{tabular}}
\caption{\textbf{Results on the \creak validation set.} Accuracies are averaged over three seeds. Results that are not significantly worse than the best result in each block are made bold. $^*$Indicates that training data includes generated statements from the Unsupervised \ourss (Imp. + Cont.) experiment along with the supervised statements.}
\label{tab:full_creak}
\end{table*}

\section{Formal Analysis of \ourss}

\newcommand{\corr}{a^*}
\newcommand{\aclm}{\ensuremath{p_{\mathrm{\ourss}}\xspace}}

At first glance, it may seem surprising that this procedure (especially in its unsupervised form) can improve LM accuracy using only LM-generated text. In this section, we describe a set of assumptions under which \ourss is \emph{guaranteed} to improve accuracy on certain inputs. We focus this analysis on generation and evaluation of (question, answer) pairs, but it could be extended to the other tasks considered in this paper as well.

Informally, suppose: 
\begin{enumerate}
    \item \emph{Questions generated by the LM with high probability are likely to be correct}. (Intuitively, high-probability questions will be ones that occurred frequently in the training set, and are therefore more likely to be answered correctly; \citealp{mccoy2023embers}, though c.f.\ \citealp{Lin2021TruthfulQAMH}.)

    \item \emph{Given a question, prompting an LM with a related, correct question--answer pair increases the probability of a correct answer}. (Intuitively, such prompts may steer models generally in the direction of truthfulness, as in \citealp{Lin2021TruthfulQAMH}, and can provide concrete evidence useful for answering the new question.)
\end{enumerate}
We wish to show that if these two conditions hold, \ourss improves model performance. 

For simplicity, we consider a minimal version of unsupervised \ourss in which a single implication is generated from each seed statement, the check in \cref{eq:consistency} is not performed, and the LM is trained to convergence on data generated from an arbitrarily large number of seeds.
Let $q$ be some specific question of interest, let $\lm(\corr \mid q)$ denote the probability that $\lm$ assigns the correct answer to $q$ (before applying \ourss), and let $\aclm(\corr \mid q)$ be the probability that the LM assigns after \ourss. Let $(q_0, a_0)$ denote a (question, answer) pair generated as a \emph{seed} document, and $\corr_0$ specifically the \emph{correct} answer to $q_0$. 
Finally, for convenience, define $p(q_0 \mid q) = \frac{\lm(q \mid q_0) \, \lm(q_0)}{\sum_{q_0'} \lm(q \mid q_0')\,  \lm(q_0')}$ (this is the probability that the seed question was $q_0$ given that the sampled question was $q$), and $p(a_0 \mid q, q_0)$ via Bayes' rule analogously.

\begin{prop}
   \label{prop:analysis}
   Suppose for some $q$ that:
   \begin{enumerate}
       \item $p(\corr_0 \mid q, q_0) \geq p^*$. (Conditioned on generating $q$ during the document generation step of \ourss, the probability that the generated answer to any seed question $q_0$ contains a correct answer is (uniformly) at least $p^*$.)

       \item $\mathbbm{E}_{q_0 \mid q} ~\lm(\corr \mid q, q_0, a_0^*) \geq \lm(\corr \mid q)\, /\, p^*$. (In expectation, conditioning on a correct $(q_0, a_0)$ pair increases the probability of generating a correct answer by at least $1/p^*$.)
   \end{enumerate}
   Then,
   \begin{equation}
       \aclm(\corr \mid q) > \lm(\corr \mid q) ~ .
   \end{equation}
   In other words, for any question $q$ satisfying the two conditions above, unsupervised \ourss increases the probability that \lm answers $q$ correctly.
\end{prop}
\noindent Proof is given in \cref{app:analysis}.
\section{Experiments}
We evaluate \ours on a set of benchmark tasks measuring fact verification, question answering with new information, and a diagnostic model editing dataset.
We use Llama-2-7B in all experiments. Additional qualitative results are provided in \cref{sec:qualitative}.

\subsection{Fact Verification} \label{sec:exp_fact_verif}

\paragraph{Task and training details}
We first evaluate whether \ourss improves the models' ability to classify factual claims.
Our experiments use \creak \cite{onoe2021creak}, a dataset of claims about entities. We investigate four different learning settings: unsupervised, supervised, semi-supervised, and transductive, each using a different procedure for sampling seed documents. We report results on the \creak development set. During \ourss fine-tuning, we use a linear learning rate schedule until the \emph{training} loss converges---this corresponds around 30 epochs for the majority of experiments unless otherwise indicated (see \cref{app:exp_details} for further details on experimental settings).

\paragraph{Evaluation and baselines} Models are scored based on the fraction of claims they correctly label as true or false. For each condition, we compare to a state-of-the-art baseline. For unsupervised \ourss, the baseline is an ordinary few-shot prompt. For supervised \ourss, the baseline fine-tunes the LM on the provided true statements. For transductive \ourss, we also compare to an inference-time baseline \textit{Graph-Inference} similar to those described by \citealp{mitchell-etal-2022-enhancing} and \citealp{kassner-etal-2023-language}, which generates implications and contradictions for each test example, performs reasoning as in \cref{eq:consistency}, then directly outputs the inferred truth value for the example (with no fine-tuning). Unlike past work, we use the base LM to generate these graphs rather than a specialized pre-trained implication generation model.
All results are presented in \cref{tab:full_creak}.

\paragraph{Results: Unsupervised \ourss} To generate seed documents, we query \lm 10 times, each time prompting the model to generate 10 diverse claims and sampling with a temperature of 0.9. We filter out the duplicate claims before continuing to sample implications and contradictions. 
The full method substantially outperforms a few-shot prompting baseline, and may outperform ablated versions of \ourss that fine-tune only on seed statements assigned a high prior probability (labeled ``seed only'' in Table~\ref{tab:creak_contrast}) or that do not perform the logical inference step described in \cref{sec:consistency-eval} (labeled ``$-$~Consistency Eval'').

For these unsupervised experiments, we perform an additional evaluation specifically aimed at measuring logical \emph{coherence} as well as factual accuracy.
Here we use the contrast set in \creak, which comprises 250 pairs of lexically similar examples with opposite truth values (e.g. \emph{Zendaya was raised in the US} and \emph{Zendaya was raised in Scotland}). In addition to accuracy, we compute the fraction of pairs that are labeled \textit{Both True} (indicating incoherence) and \textit{Both Correct}. 

Here, \ourss not only improves correctness but also reduces the number of incoherent predictions, decreasing the probability that \lm judges two contradictory statements to both be correct.

\begin{table}
\centering
\resizebox{0.48\textwidth}{!}{
\begin{tabular}{lrrr}
\toprule
\textbf{Method} &  \textbf{Both True \textdownarrow }&  \textbf{Both Correct \textuparrow }&  \bf Acc. \textuparrow \\
\midrule
\bf Prompting              &       34.4 &          36.8 &      63.2 \\
\bf \ourss (Seed only)             &       20.4 &          47.6 &      72.2 \\
\bf \ourss (Cont.)       &       \textbf{12.0} &          47.6 &      72.0 \\
\bf \ourss (Imp. + Cont.) &       19.2 &          \textbf{49.6} &      \textbf{73.0} \\
\bottomrule
\end{tabular}}
\caption{\textbf{Logical coherence (Both True) and factuality (Both Correct) for unsupervised \ourss 
 on the \creak contrast set.} \ourss not only increases accuracy, but decreases the number of logically incoherent predictions (in which \lm assigns labels two contradictory statements as both true). \label{tab:creak_contrast}}
\end{table}
\begin{table*}
\centering
\resizebox{0.82\textwidth}{!}{
\begin{tabular}{lccccc}
\toprule
\bf Method & \multicolumn{5}{c}{\bf Number of Edits} \\
&   \bf 10  &  \bf 20 & \bf 50 & \bf 100 & \bf 1000  \\
\midrule
\textit{Retrieval-based} \\
\bf MeLLo \cite{zhong-etal-2023-mquake} & 15.3 \stderr{4.4} & 18.3 \stderr{3.5} & 12.0 \stderr{1.0} & 12.1 \stderr{0.7} & 11.4 \\
\midrule
\textit{Parameter-update-based} \\
\bf Fine-tuning on Edits                             &   0.7 \stderr{1.5} &   9.0  \stderr{5.4} &   5.5 \stderr{1.8} &   6.7 \stderr{1.2} &   4.1 \\
\bf FT on Continuations \cite{padmanabhan2023propagating} & 4.4 \stderr{2.9} & 4.4 \stderr{2.0} & 3.1 \stderr{1.2} & 3.6 \stderr{1.2} & 3.3\\
\bf MEMIT \cite{meng2022memit} & 11.1 \stderr{2.9} & 11.7 \stderr{5.1} & 6.0 \stderr{0.7} & 1.1 \stderr{0.6} & 0.6\\
 \bf \ourss (Imp.)                           &  20.0 \stderr{6.7} &  20.5 \stderr{6.1} &  14.0 \stderr{5.2} &  10.7 \stderr{1.9} &   8.1 \\
 \bf \ourss (Corr. Imp.)           &  \textbf{41.7} \stderr{5.0} & \textbf{29.4} \stderr{6.8} &  \textbf{25.6} \stderr{3.8} & \textbf{14.2} \stderr{1.2} &  12.3 \\
\bf \ourss (Corr. Imp. + Imp.) &  \textbf{30.0} \stderr{6.7} &  \textbf{35.6} \stderr{4.8} &  \textbf{15.6} \stderr{7.3} &  \textbf{18.9} \stderr{2.1} &  \textbf{15.4} \\ 
\bottomrule
\end{tabular}}
\caption{\textbf{\mquake counterfactual subset results.} We provide average test set accuracy (standard errors are given in parentheses) across three seeds except for 1,000 where we evaluate only once. Results that are not significantly different from the best score are made \textbf{bold} (paired $t$-test $p \ll 0.05$). For each edit, there are 3 multi-hop test questions. Before fine-tuning we convert each edit into a question using prompting. In \ourss (Corr. Imp.), we prompt the model to first produce related facts to the initial claim before generating implications.
}
\label{tab:mquake}
\end{table*}

\paragraph{Results: Supervised \& Semi-supervised \ourss} In the supervised case (\cref{tab:full_creak}), we utilize a small set of externally provided claims and associated ground-truth labels to initialize \ourss seed nodes. We sample 20 claims from the \creak training set and filter those labeled as \texttt{true} to use as our seed documents $D$. For semi-supervised learning, we pool together data generated following the unsupervised and supervised settings for fine-tuning. 

All variants of \ourss improve over an ordinary fine-tuning baseline; interestingly, examples generated supervisedly and self-supervisedly are complementary, such that semi-supervised learning improves over both results.

\paragraph{Results: Transductive \ourss} 
The previous evaluations assumed a strict train / test split. Here we study the behavior of DCT in a ``transductive'' setting \citep{learn2013gammer} in which we have access to \emph{unlabeled} claims from the evaluation set while updating the model. For each of the 1,371 claims in the validation set, we generate seed text by prompting the LM to generate a set of \emph{related} claims, which are then used to generate additional implications and contradictions. In addition to the inference-time baseline described above, these experiments compare to an ablated version of DCT that trains only on the generated related claims.

As in other experiments, \ourss outperforms the inference-time reasoning baseline as well as the related-text-only ablation.

\subsection{Model Updating and Question Answering}

\paragraph{Task and training details}

Language models often hallucinate wrong information and rapidly become out-of-date after initial training. As a consequence, there has been increased interest in specialized continual learning (or ``model editing'') procedures for updating LMs with new information without full re-training. A key desideratum is LMs should not simply assign high probability to the new fact, but all of its \emph{consequences}:
if we wish to update an LM encode the fact that the current U.K.\ prime minister is not Boris Johnson but Rishi Sunak, the LM should also produce text consistent with the fact that the current P.M.'s wife is not Carrie Johnson but Akshata Murthy. Past work has found that fine-tuning on edits, as well as many specialized editing procedures, fail to propagate such information.

Our experiments on this task use the counterfactual subset from \mquake \cite{zhong-etal-2023-mquake} dataset, which evaluates models on their ability to answer questions about new information not provided in their training sets.
To apply \ourss, we take as seed documents
the text of the new information to be inserted into the model. During the
generation phase, models are prompted to combine this information with other
background knowledge related to the same topic (see \cref{sec:prompt_templates} for prompting
details), producing what we term \textit{Correlative Implications}. 
Finally, because \mquake is a question answering dataset, we convert each generated statement into a question--answer pair 
using the LM, then fine-tune it on these pairs.

\paragraph{Evaluation and baselines}
We compare \ourss to ordinary fine-tuning on new information and three state-of-the-art baseline approaches for model updating: a context distillation baseline by \citet{padmanabhan2023propagating}, which fine-tunes LMs to behave out-of-context the same way they would with prompts containing the new information (see \cref{app:exp_details} for implementation details), a weight editing baseline by \citep{meng2022memit}, and the retrieval baseline
MeLLo \cite{zhong-etal-2023-mquake}, which stores new text in an external memory. 
We evaluate the behavior of \ourss and these baselines in settings where varying numbers of new pieces of information (between 10 and 1000) are provided, and report the model's accuracy at question answering.

\paragraph{Results} As shown in \cref{tab:mquake}, \ourss significantly outperforms fine-tuning, fine-tuning on continuations, weight editing, and MeLLo (the previous state-of-the-art on \mquake). Using correlative implications systematically improves over simple implications. Combining the two sets improves on average over using either in all settings. Our qualitative analysis in \cref{sec:qualitative} reveals that correlative implications contain about 50\% more new information than standard implications.
 
\subsection{Sanity Checks for LM Consistency}

\paragraph{Task and training details}
In addition to naturalistic question asking tasks like \mquake, there has been recent interest in developing precise tests of LMs' ability to capture simple logical implications of new facts (e.g.\ assigning high probability to sentences of the form \emph{B is A} after training on \emph{A is B}). We investigate whether \ourss can address these issues using the \reversal benchmark \citep{berglund2023reversal}.
We report results on two evaluations: first, a set of celebrity parent--child pairs with training examples like \textit{Jennifer Lawrence's mother is Karen Lawrence} and test examples \textit{Who is the child of Karen Lawrence?}; second, a set of entity--description pairs with training examples like \textit{Olaf Scholz was the ninth Chancellor of Germany} and cloze-style test examples \textit{The ninth Chancellor of Germany is \underline{\hspace{1em}}}.

\paragraph{Evaluation and baselines}
For these experiments, we compare to the fine-tuning baseline used in the original work of \citet{berglund2023reversal} as well as the fine-tuning on continuations approach by \citet{padmanabhan2023propagating}. We use training examples as seed statements, and generate implications using \emph{the same prompt as \creak experiments in \ref{sec:exp_fact_verif}}. While we expect that a \ourss-type approach specifically tailored for this benchmark could trivially re-generate all the test examples, our experiments in this section aim to evaluate whether a general-purpose prompt can improve performance on a specific class of generalizations. Following \citet{berglund2023reversal}, we report exact-match accuracy after removing punctuation and lower-casing.
In this dataset, LMs are evaluated on a mix of questions and cloze completion tasks
featuring both training statements and their reversed forms.

\paragraph{Results}

Results are shown in \cref{tab:reversal_all}. \ourss 
improves accuracy on reversed statements without significantly hurting performance on original questions. Notably, however, \ourss with this general-purpose prompt does not completely solve this dataset, and we leave for future work the question of whether more extensive sampling or other procedures could further improve these results.

\begin{table}
\centering
\resizebox{0.45\textwidth}{!}{
\begin{tabular}{lrrr}
\toprule
 & \multicolumn{2}{c}{Direction} \\
 & \textbf{Same} & \textbf{Reverse}  & \textbf{Average}\\
\midrule
\textit{Child-to-Parent} \\
\textbf{Fine-tuning} &   \textbf{95.3} &   2.2  & 48.7\\
\textbf{FT \cite{padmanabhan2023propagating}} & 57.3 & 7.1 & 32.2\\
\textbf{\ourss (Imp.)} & 87.9 & \textbf{48.3} & \textbf{68.1} \\ \midrule
\textit{Person-to-Description}\\
\textbf{Fine-tuning} &   \textbf{83.7} &                3.7 &  43.7 \\
\textbf{FT \cite{padmanabhan2023propagating}} & 54.3 & \textbf{27.0} & 40.7\\
\textbf{\ourss (Imp.)} & 81.3 & 10.7 & \textbf{46.0} \\ \midrule
\textit{Description-to-Person}\\
\textbf{Fine-tuning} & \textbf{99.7} &                3.0 &  51.3 \\
\textbf{FT \cite{padmanabhan2023propagating}} & 99.3 & 1.0 & 50.2 \\
\textbf{\ourss (Imp.)} & \textbf{99.7} &                \textbf{15.7} &  \textbf{57.7} \\
\bottomrule
\end{tabular}}
\caption{\textbf{Reversal Curse benchmark results.} While this challenge remains far from solved, applying \ourss (with the same prompt used for \creak experiments) substantially improves accuracy. We use 1,000 examples for \textit{Child-To-Parent} and 300 for the other two subsets for evaluation.}
\label{tab:reversal_all}
\end{table}

\section{Conclusion}
We have described \ours (\ourss), a supervision procedure that optimizes models toward deductive closure---encouraging them to assign high probability to a logically coherent set of factual assertions.
By doing so, \ourss also improves the truthfulness and updatability of models, substantially increasing accuracy on a variety of fact verification and editing datasets in both supervised and unsupervised conditions. More generally, these results show that some factual errors in LMs stem not from limitations of their training data, but limitations of training algorithms. By using LMs themselves to reason about relationships between (and implications of) their predictions, they can be made more accurate with little or no additional supervision.

\section*{Limitations}
While Deductive Closure Training (DCT) could in principle be applied to arbitrary graphs of relations between statements, here we have applied it only to a single layer of implications of seed data. All datasets used for evaluation involve English text, and it is possible that DCT behaves differently in different languages. Even within English, it is possible that exhibits systematic biases or differences in accuracy for certain types of factual content. While DCT can improve overall factuality, it may inadvertently perpetuate hallucinations within certain domains that could escape detection during our evaluations.

\section*{Ethical Considerations}
While our experiments have focused on using \ourss as a tool for bringing LMs into alignment with reliable sources, these techniques could also be used to optimize LMs toward generation of (logically consistent) false facts, increasing their effectiveness as tools for generation of misinformation.

\section*{Acknowledgments}
This work was supported partly by the National Science foundation under grant IIS-2238240, DARPA HR001118S0044 (the LwLL program), the Shared Computing Cluster administered by Boston University's Research Computing Services, as well as a hardware donation from NVIDIA to MIT. Any opinions, findings, conclusions, or recommendations expressed here are those of the authors and do not necessarily reflect the view of the sponsor.

\bibliography{anthology,custom}

\appendix

\clearpage
\section{Experimental Details}
\label{app:exp_details}
We use the \texttt{Llama-2-7B-hf} checkpoint provided by HuggingFace Transformers library for all of our experiments. Code to reproduce the experiments will be made publicly available. While developing the codebase, the authors used GitHub Copilot via Visual Studio Code.

\paragraph{Generation} We sample at temperature 0.6 and top-p 0.9 for all samples except for the set of seed documents for the unsupervised experiment in \cref{tab:full_creak} where we used temperature 0.9 to obtain a diverse set of initial documents.
\paragraph{Training} For fine-tuning we use the LoRA implemention via the PEFT library \cite{hu2022lora,peft} and set rank to 8, alpha to 32 and dropout to 0.1. In the absence of a held-out development set, we set the learning rate to 0.0001 throughout, batch size to 4 and train for 30 epochs by default. We find that training loss typically converges after 30 epochs with the exception of the supervised experiments in \cref{tab:full_creak} for which we train for 60 epochs. The transductive setting for \creak results in substantially more training documents, hence we train only for 1 epoch. We use a linear learning rate scheduler with 100 warm up steps and AdamW optimizer. For fact verification training, we use weighted sampling as the class distribution is sometimes unbalanced.

\paragraph{Editing experiments} We use the \mquake-CF subset from \citet{zhong-etal-2023-mquake} and evaluate only on the multi-hop questions. \citet{padmanabhan2023propagating} proposes two techniques to introduce model updates based on fine-tuning: simple fine-tuning on continuations conditioned on the edit statement (which we call \textit{FT on Continuations}) and context distillation on continuations. We find the former approach--fine-tuning the model on the continuations when the model is conditioned on the edit sequence--to perform better on \mquake than the latter. Hyperparameters used for MEMIT are available in \cref{tab:memit_hyperparams}. For validation we use a set of held-out 50 edits.

\begin{table}
\centering
\label{tab:memit_hyperparams}
\resizebox{0.3\textwidth}{!}{%
\begin{tabular}{ll}
\toprule
Parameter & Value \\ \midrule
\texttt{layers} & [3, 4, 5, 6, 7] \\
\texttt{clamp\_norm\_factor} & 4.0 \\
\texttt{layer\_selection} & all \\
\texttt{fact\_token} & subject\_last \\
\texttt{v\_num\_grad\_steps} & 25 \\
\texttt{v\_lr} & 5e-1 \\
\texttt{v\_loss\_layer} & 31 \\
\texttt{v\_weight\_decay} & 0.5 \\
\texttt{kl\_factor} & 0.0625 \\
\texttt{mom2\_adjustment} & true \\
\texttt{mom2\_update\_weight} & 15000 \\
\texttt{mom2\_dataset} & wikipedia \\
\texttt{mom2\_n\_samples} & 100000 \\
\texttt{mom2\_dtype} & float32 \\
\bottomrule
\end{tabular}
}
\caption{MEMIT hyperparameters.}
\end{table}

\section{Details for DCT}
In \cref{tab:sample_consistency}, we consider a small graph consisting of one seed node ($r_i$), one implication ($r_{i1}$) and one contradiction ($r_{i2}$). In the beginning, there are 8 candidate truth value assignment yet not all assignments are consistent within e.g. If $r_i$ if true, then $r_{i1}$ must be true and $r_{i2}$ must be false. When computing the most probable assignment in \cref{eq:consistency}, we only consider consistent assignments.

\begin{table}[]
\resizebox{0.46\textwidth}{!}{%
\begin{tabular}{cccc}
\toprule
\multicolumn{3}{c}{Truth Value Assignment ($T_i$)} \\ \midrule
Seed & Implication & Contradiction & Consistency $c(T_i)$ \\ \midrule
T    & T           & T   & 0          \\
T    & T           & F   & 1          \\
T    & F           & T   & 0          \\
T    & F           & F   & 0          \\
F    & T           & T   & 1          \\
F    & T           & F   & 1          \\
F    & F           & T   & 1          \\
F    & F           & F   & 1         \\ \bottomrule
\end{tabular}%
}
\caption{Consistency evaluations candidate truth value assignments for a small graph of three nodes: one seed, one implication and one contradiction documents.}
\label{tab:sample_consistency}
\end{table}

\section{Qualitative Analysis}

\label{sec:qualitative}
To better understand how \ourss improves LM performance,
we manually annotated about 350 generations from various experiments to assess whether (1) double-checking improves the precision of generated implications and contradictions; (2) whether \ourss incorporates model internal knowledge when making new conclusions; and (3) whether generated text includes non-trivial new inferences.

\paragraph{Double-checking} We evaluated whether the double-checking following \ourss (Imp. + Cont.) improves precision. In the supervised setting for \creak, we annotated 100 implications and contradictions generated using \ourss (Imp. + Cont.). We found that 74 of these are valid. The double-checking procedure removes about 2/3 of generations, resulting in 33. Among these, 27 are valid, raising the ratio of correct statements predicted by the model from 76\% to 82\%.

\paragraph{Incorporating previous information} The \mquake subset used in our experiments comprises difficult multi-hop questions. Hence, generations that incorporate existing information about the entities mentioned in the edit are especially useful. We compare the set of implications generated using the \ourss (Imp.) and \ourss (Corr. Imp.). 
Respectively, only 30\% and 36\% of generations involve strict logical implications; however, 
78\% and 69\% were judged to be plausible given the edit. Furthermore, 24\% and 33\% of the generations incorporate new information supplied by the LM. For example, given an edit \textit{Chauncey Billups is associated with the sport of pesäpallo}, the LM uses background knowledge \textit{Pesäpallo is popular in Finland} to generate \textit{Chauncey Billups was born in Finland}.

\paragraph{Novelty of inferences} Lastly, we find that most implications made by the model on the \reversal dataset are paraphrases or are trivial (\emph{Jennifer Lawrence's mother is Karen Lawrence} $\to$ \emph{Jennifer Lawrence has a mother}) but some add world knowledge to the implication (\emph{Sadie Frost's mother is Mary Davidson} $\to$ \emph{Mary Davidson is the mother of a British actress}, where the LM itself has supplied the knowledge about Sadie Frost). While generating implications, \ourss often (but not always) generates test-set-like reversed implications on its own: the model reverses 22\% of the statements of the form \emph{X's parent is Y}, 43\% of statements of the form \emph{the person with property X is Y}, but only 6\% of statements of the form \textit{Person X has property Y}. These findings suggest a strong bias toward generating text that starts with the person as opposed to the description. In general, most generated extensions are fluent, different from the source, and sometimes contain new information.



\section{Prompt Templates}
\label{sec:prompt_templates}
We use a set of fixed prompts to generate our graphs, calculate model-estimated probability for the correctness of a given statement, generating a set of seed documents and automatically converting statements into questions which are available in \cref{tab:prompt_templates,tab:double-check-prompts,tab:similar_unsup_prompts,tab:qa_conversion}.

\onecolumn
\begin{table}[h]
\caption{\textbf{Implication \& contradiction prompt templates.}}
\centering
\resizebox{0.96\textwidth}{!}{
\begin{tabular}{p{5.5cm}p{14cm}}
\toprule
Procedure & Prompt \\
\midrule
Implication & \texttt{List three implications of the given claims.}

\texttt{Claim: Cleopatra was the last active ruler of the Ptolemaic Kingdom of Egypt between 51 to 30 BC.}\newline
\texttt{Logical implications:}\newline
\texttt{1. Cleopatra was one of the rulers of the Ptolemaic Kingdom of Egypt.}\newline
\texttt{2. Egypt had a female ruler during the Ptolemaic Kingdom age.}\newline
\texttt{3. Ptolemaic Kingdom of Egypt ended on 30 BC.}\newline

\texttt{Claim: \{claim\}}\newline
\texttt{Logical implications:}\newline\\
Implication (\mquake) & \texttt{List five logical implications of the given claims.}\newline

\texttt{Claim: Stephen Hawking was born and raised in Russia.}\newline
\texttt{Logical implications:}\newline
\texttt{1. Stephen Hawking has knowledge of Russian language.}\newline
\texttt{2. The head of the country where Stephen Hawking was born is Vladimir Putin.}\newline
\texttt{3. The country where Stephen Hawking was born is Russia.}\newline
\texttt{4. Stephen Hawking is a Russian citizen and has a Russian passport.}\newline
\texttt{5. The city where Stephen Hawking was born is in Russia.}\newline

\texttt{Claim: \{claim\}}\newline
\texttt{Logical implications:}\newline\\
Correlative Implication (\mquake) & \texttt{Given a main claim, list five related facts, and then logical implications of the claim and related fact.}\newline

\texttt{Main Claim: Stephen Hawking was born and raised in Russia.}\newline
\texttt{Related Facts:}\newline
\texttt{1. The language of Russia is Russian.}\newline
\texttt{2. The head of Russia is Vladimir Putin.}\newline
\texttt{3. Russia is on the continents of Asia and Europe.}\newline
\texttt{4. The capital of Russia is Moscow.}\newline
\texttt{5. The currency of Russia is Russian ruble.}\newline

\texttt{Implications:}\newline
\texttt{1. Stephen Hawking has knowledge of Russian language.}\newline
\texttt{2. The head of the country where Stephen Hawking was born is Vladimir Putin.}\newline
\texttt{3. The country where Stephen Hawking was born is on the continents of Europe and Asia.}\newline
\texttt{4. The capital of Stephen Hawking's home country is Moscow.}\newline
\texttt{5. Stephen Hawking has used Russian ruble growing up.}\newline

\texttt{Main Claim: \{claim\}}\newline
\texttt{Related Facts:}\newline\\
\bottomrule
\end{tabular}}
\label{tab:prompt_templates}
\end{table}

\begin{table}[h!]
\caption{\textbf{Prompt templates for double-checking, generating similar claims and estimating model-assigned truth value.}}
\centering
\resizebox{0.96\textwidth}{!}{
\begin{tabular}{p{5cm}p{14cm}}
\toprule
Procedure & Prompt \\
\midrule
Implication (Double-Check) & \texttt{For the given pair of claims you need to decide if the first one implies the second. Give your final verdict at the end. Here are some examples.}\newline

\texttt{The tallest building in the world is taller than 800 metres.}\newline
\texttt{The tallest building in the world is taller than 700 metres.}\newline
\texttt{Discussion: If something is taller than 800 then it is necessarily taller than 700.}\newline
\texttt{Final Verdict: Implies.}\newline

\texttt{Orange is a fruit.}\newline
\texttt{Orange is an apple.}\newline
\texttt{Discussion: Not all fruit are apples so orange being a fruit does not imply that is also an apple.}\newline
\texttt{Final Verdict: Does not imply.}\newline

\texttt{\{claim1\}}\newline
\texttt{\{claim2\}}\newline
\texttt{Discussion:}\newline\\
Contradiction (Double-Check) & \texttt{For the given pair of claims you need to decide if they are contradictory or not. Give final verdict at the end. Here are some examples.}\newline

\texttt{Claim 1: The tallest building in the world is taller than 800 metres.}\newline
\texttt{Claim 2: The tallest building in the world is shorter than 1000 metres.}\newline
\texttt{Reasoning: A building can be both taller than 800 and shorter than 1000.}\newline
\texttt{Final Verdict: Not contradictory.}\newline

\texttt{Claim 1: Orange is a fruit.}\newline
\texttt{Claim 2: Orange is a vegetable.}\newline
\texttt{Reasoning: Fruit and vegetable are disjoint categories.}\newline
\texttt{Final Verdict: Contradictory.}\newline

\texttt{Claim 1: \{claim1\}}\newline
\texttt{Claim 2: \{claim2\}}\newline
\texttt{Reasoning:}\newline\\

Estimating Truth Value & \texttt{Label the following statements according to whether or not they are true:}\newline
\texttt{World War II began in 1965. Label: false}\newline
\texttt{Alan Alda is an actor. Label: true}\newline
\texttt{The moon is made of obsidian. Label: false}\newline
\texttt{There are approximately 30 million people in the United States. Label: false}\newline
\texttt{Dracula was written by Bram Stoker. Label: true}\newline
\texttt{\{claim\} Label:}\newline\\
\bottomrule
\end{tabular}}
\label{tab:double-check-prompts}
\end{table}

\begin{table}[h!]
\caption{\textbf{Prompt templates for generating contradictions, related statements (used in the transductive setting) and unsupervised seed document generation.}}
\centering
\resizebox{0.96\textwidth}{!}{
\begin{tabular}{p{5cm}p{14cm}}
\toprule
Procedure & Prompt \\
\midrule
Contradiction & \texttt{Given a claim, generate three other very similar-looking but CONTRADICTING claims.}\newline

\texttt{Claim: Cleopatra was the last active ruler of the Ptolemaic Kingdom of Egypt between 51 to 30 BC.}\newline
\texttt{Similar but contradicting claims:}\newline
\texttt{1. Cleopatra was the first active ruler of the Ptolemaic Kingdom of Egypt.}\newline
\texttt{2. Cleopatra was the last active ruler of the Ptolemaic Kingdom of Egypt between 51 to 30 AD.}\newline
\texttt{3. Cleopatra was the daughter of the last active ruler of the Ptolemaic Kingdom of Egypt.}\newline

\texttt{Claim: \{claim\}}\newline
\texttt{Similar but contradicting claims:}\\
Similar claims ($\relprompt$)& \texttt{Generate five related factual statements on the same topic as the given claim. Note that the given claim may or may not be correct. However, the generated statements should each be correct and different.}\newline
\texttt{Claim (may be true or false): Neil Armstrong and Buzz Aldrin became the first humans to land on the Mars.}\newline
\texttt{Related Correct Facts:}\newline
\texttt{1. Apollo 11 was the first manned mission to land on the moon.}\newline
\texttt{2. Neil Armstrong was the first person to step on the moon.}\newline
\texttt{3. No human has been to Mars yet.}\newline
\texttt{4. Neil Armstrong and Buzz Aldrin were the first humans to land on the moon.}\newline
\texttt{5. Neil Armstrong and Buzz Aldrin were the first humans to walk on the moon.}\newline
\texttt{Claim (may be true or false): \{claim\}}\newline
\texttt{Related Correct Facts:}\newline\\
Unsupervised seed claims & \texttt{Generate ten examples of factual claims. List your claims in separate lines.}\newline
\texttt{1.}\newline\\
\bottomrule
\end{tabular}}
\label{tab:similar_unsup_prompts}
\end{table}

\begin{table}[h!]
\caption{\textbf{Prompt template for converting model-generated statements into questions. We re-use the original statements as the corresponding answers.}}
\centering
\resizebox{0.96\textwidth}{!}{
\begin{tabular}{p{5cm}p{14cm}}
\toprule
Procedure & Prompt \\
\midrule
Conversion to questions & \texttt{Sentence: Kate Winslet is a citizen of the UK.}\newline
\texttt{Question: Which country is Kate Winslet a citizen of?}\newline
\texttt{Sentence: Ukraine is a country in Europe.}\newline
\texttt{Question: Which continent is Ukraine in?}\newline
\texttt{Sentence: The country where Priyanka Chopra is from is India. The capital of India is New Delhi.}\newline
\texttt{Question: What is the capital of the country where Priyanka Chopra is from?}\newline
\texttt{Sentence: {sentence}}\newline
\texttt{Question:}\\
\bottomrule
\end{tabular}}
\label{tab:qa_conversion}
\end{table}

\clearpage
\section{Proof of \cref{prop:analysis}}
\label{app:analysis}

At optimality, $\aclm(\corr \mid q)$ (the probability that the updated LM assigns to the correct answer) will be the probability of $\corr$ given $q$ marginally over all generated seed documents:
\begin{align*}
    \aclm(\corr \mid q) &= \sum_{q_0, a_0} \lm(\corr \mid q, q_0, a_0) \, p(a_0 \mid q_0, q) \, p(q_0 \mid q) ~ .
\intertext{We may decompose this according to whether the generated seed pair is itself correct:}
    &= \sum_{q_0} p(q_0 \mid q) \Big[ \lm(\corr \mid q, q_0, \corr_0) \, p(\corr_0 \mid q, q_0) \\
    & \hspace{7em} + \sum_{a_0' \neq \corr_0} \lm(\corr \mid q, q_0, a_0') \, p(a_0' \mid q_0, q) \, p(q_0 \mid q)  \Big] \\
\intertext{(where $\corr_0$ denotes the correct answer to $q_0$)}
    &\geq \sum_{q_0} p(q_0 \mid q) \, \lm(\corr \mid q, q_0, \corr_0) \, p(\corr_0 \mid q, q_0) ~ .\\
\intertext{By assumption 1:}
    &\geq \sum_{q_0} p(q_0 \mid q) \lm(\corr \mid q, q_0, \corr_0) \, p^* \\
    &= p^* \, \mathbbm{E}_{q_0 \mid q} ~ \lm(\corr \mid q, q_0, \corr_0) ~ . \\
\intertext{By assumption 2:}
    &\geq \lm(\corr \mid q) ~ . \quad \qedsymbol
\end{align*}

\end{document}